\documentclass[11pt]{article}

\usepackage[final]{acl}

\usepackage{times}
\usepackage{latexsym}
\usepackage[T1]{fontenc}
\usepackage[utf8]{inputenc}
\usepackage{microtype}
\usepackage{inconsolata}
\usepackage{graphicx}

\usepackage[ruled,vlined]{algorithm2e}
\usepackage{booktabs}
\usepackage{multirow}
\usepackage{amsmath}
\usepackage{pdfpages}
\usepackage{afterpage}
\usepackage{stfloats}
\usepackage{amsmath}
\usepackage{xspace}
\usepackage{cleveref}
\usepackage[normalem]{ulem}
\useunder{\uline}{\ul}{}
\usepackage{enumitem}
\usepackage[table]{xcolor}
\usepackage{colortbl}
\usepackage{amssymb}
\usepackage{makecell}
\usepackage{xfp}

\usepackage[most]{tcolorbox}
\tcbuselibrary{breakable} 
\usepackage{multicol}    
\usepackage{calc}      
\newsavebox{\autocolbox}
\newcommand{\AutoCols}[2][0.32\textheight]{%
  \sbox{\autocolbox}{\begin{minipage}{\linewidth}#2\end{minipage}}%
  \ifdim\ht\autocolbox>#1
    \begin{multicols}{2}
      \setlength{\columnsep}{10pt}
      #2
    \end{multicols}%
  \else
    #2%
  \fi
}

\usepackage{listings} 
\lstdefinelanguage{yaml}{
  keywords={true,false,null,y,n},
  sensitive=false,
  comment=[l]{\#},
  morestring=[b]',
  morestring=[b]"
}

\definecolor{valgreen}{RGB}{46,139,87}
\definecolor{valred}{RGB}{178,34,34}
\newcommand{\ERRmax}{0.5}
\newcommand{\ERR}[1]{%
  \begingroup
  \edef\colorpct{\fpeval{round(20 + 60*min(abs(#1)/\ERRmax,1),0)}}
  \ifdim #1 pt > 0pt
    \edef\colorspec{valgreen!\colorpct}%
  \else
    \edef\colorspec{valred!\colorpct}%
  \fi  \expandafter\cellcolor\expandafter{\colorspec}%
  #1%
  \endgroup
}

\newcommand{\TaskName}{\emph{calendar conflict resolution}}
\newcommand{\BenchName}{\textsc{CalConflictBench}}
\newcommand{\ModelName}{\textbf{\textcolor{darkgray}{\textsc{PEARL}}}}

\NewDocumentCommand{\heng}
{ mO{} }{\textcolor{red}{\textsuperscript{\textit{Heng}}\textsf{\textbf{\small[#1]}}}}
\NewDocumentCommand{\xiusi}
{ mO{} }{\textcolor{cyan}{\textsuperscript{\textit{Xiusi}}\textsf{\textbf{\small[#1]}}}}
\NewDocumentCommand{\jeongh}
{ mO{} }{\textcolor{orange}{\textsuperscript{\textit{Jeonghwan}}\textsf{\textbf{\small[#1]}}}}
\NewDocumentCommand{\cheng}
{ mO{} }{\textcolor{orange}{\textsuperscript{\textit{Cheng}}\textsf{\textbf{\small[#1]}}}}

\usepackage{todonotes}

\title{\ModelName{}: Self-Evolving Assistant for Time Management \\ with Reinforcement Learning}

\newcommand{\Authorlist}{
Bingxuan Li$^1$,~~Jeonghwan Kim$^1$,~~Cheng Qian$^1$, ~~Xiusi Chen$^1$,  \\ Eitan Anzenberg$^2$,~~Niran Kundapur$^2$,~~Heng Ji$^1$
}
\newcommand{\AuthorAffliations}{
$^1$ University of Illinois at Urbana-Champaign \quad
$^2$ Eightfold.ai
}
\newcommand{\ReleaseDate}{January, 17th, 2026}
\newcommand{\Contact}{bl61@illinois.edu, hengji@illinois.edu}
\newcommand{\ProjectWebsite}{https://bx126.github.io/pearl.github.io}

\usepackage{tcolorbox}
\tcbuselibrary{skins,breakable}

\definecolor{brandblue}{RGB}{51,102,0}
\definecolor{brandline}{RGB}{51,102,0}

\newtcolorbox{StyledAbstractBox}{
  enhanced,
  breakable,
  colback=white,
  colframe=brandline,
  boxrule=1.5pt,
  arc=4mm,
  left=14pt,right=14pt,top=12pt,bottom=12pt,
}

\makeatletter
\newcommand{\IfEmptyTF}[3]{%
  \if\relax\detokenize{#1}\relax #2\else #3\fi
}

\newcommand{\makestyledfirstpage}{%
  \twocolumn[%
    \thispagestyle{empty}

    \noindent{\color{brandline}\rule{\textwidth}{2pt}}
    \vspace{0.25cm}

    \begin{center}
      {\LARGE\bfseries \@title \par}
    \end{center}

    \vspace{0.25cm}
   \noindent{\color{brandline}\rule{\textwidth}{2pt}}

    \vspace{0.55cm}
    \begin{center}
      {\large \textbf{\Authorlist} \par}
      \vspace{0.2cm}
      {\large \AuthorAffliations \par}
    \end{center}

    \vspace{0.5cm}
    \begin{StyledAbstractBox}
      \begin{center}
        {\Large\bfseries\color{brandblue} Abstract\par}
      \end{center}
      \vspace{0.25cm}

      Overlapping calendar invitations force busy professionals to repeatedly decide which meetings to attend, reschedule, or decline. We refer to this preference-driven decision process as calendar conflict resolution. Automating this decision process is crucial yet challenging. Scheduling logistics can drain hours, and human delegation often fails at scale, which motivates us to ask: Can we trust large language models (LLMs) or language agents to manage time? 
To enable a systematic study of this question, we introduce \BenchName{}, a benchmark for long-horizon calendar conflict resolution. In \BenchName{}, conflicts are presented to agents round-by-round over a calendar year, requiring them to infer and adapt to user preferences progressively. Our experiments show that current LLM agents perform poorly with high error rates, e.g., Qwen-3-30B-Think has an average error rate of 35\%. To address this gap, we propose \ModelName{}, a reinforcement-learning framework that (i) augments the language agent with an external preference memory that stores and updates inferred strategies (e.g., attendee priorities, topic importance, time/location preferences), and (ii) optimizes the agent with round-wise rewards that directly supervise decision correctness, ranking quality, and memory usage across rounds. Experiments on \BenchName{} show that \ModelName{} achieves an error reduction rate of 0.76 and a 55\% improvement in average error rate compared to the strongest baseline.

      \vspace{0.1cm}
      \noindent{\color{brandline}\rule{\linewidth}{0.5pt}}
      \vspace{0.1cm}
      
      {
      \noindent
      \IfEmptyTF{\ReleaseDate}{}{%
        \textcolor{brandblue}{{\bfseries Date: }}\ReleaseDate
        \vspace{2pt}
      }%
      \\
      \IfEmptyTF{\Contact}{}{%
        \hspace{1.2em}{\bfseries \textcolor{brandblue}{Contact:}} \texttt{\Contact}
        \vspace{2pt}
      }%
      \\
      \IfEmptyTF{\ProjectWebsite}{}{%
        \hspace{1.2em}\textcolor{brandblue}{{\bfseries Project Webpage: }} \texttt{\ProjectWebsite}
      }%
      \par
      }
    \end{StyledAbstractBox}

    \vspace{0.6cm}
  ]
}
\makeatother

\title{\ModelName{}: Self-Evolving Assistant for Time Management \\ with Reinforcement Learning}

\author{
Bingxuan Li$^1$,~~Jeonghwan Kim$^1$,~~Cheng Qian$^1$, ~~Xiusi Chen$^1$,  \\ \textbf{Eitan Anzenberg$^3$},~~\textbf{Niran Kundapur$^2$},~~\textbf{Heng Ji}$^1$\\
$^1$ University of Illinois at Urbana-Champaign \quad
$^2$ Viven 
\quad
$^3$ Eightfold.ai \\
\texttt{\{bl61, hengji\}@illinois.edu}\\\\
\href{https://bx126.github.io/pearl.github.io}{\textcolor{teal}{\texttt{https://bx126.github.io/pearl.github.io}}}
}

\begin{document}
\maketitle

\section{Introduction}
\label{sec:intro}



Receiving overlapping calendar invitations is common in modern workplaces. Consider a CEO of a company or a PI of a research lab, they need to coordinate a large amount of events with different stakeholders every day, but their daily working hours are limited. When multiple events conflict with each other, they must decide which event to attend, which to postpone, and which to decline. We refer to this repeated, preference-driven decision problem as \emph{calendar conflict resolution}.

Automating calendar conflict resolution is important because it quietly drains time and undermines productivity. Scheduling logistics associated with meetings, e.g., coordinating availability or rescheduling around last-minute conflicts, can easily amount to hours each week. Workplace statistics suggest that 43\% of professionals spend at least three hours per week on scheduling meetings \citep{reclaim_smart_meetings_2024, calendly_automated_scheduling_2024, microsoft_infinite_workday_2025}. While in practice these decisions are often delegated to human assistants such as administrative staff \citep{bls_secretaries}, it can easily break down at scale. Not only do human assistants frequently confront a high volume of tasks, but they also must coordinate multiple stakeholders' schedule in order to reliably resolve scheduling logistics. Furthermore, when a conflict occurs, human assistants have to rely on sparse, incomplete signals about what the delegator values to resolve the conflict. This causes their internal preference model to drift over time, leading to judgments that are distant from the delegator's preferences.
This calls for a reliable agent that can resolve calendar conflicts.
Concretely, a reliable calendar conflict resolution agent should:
(i) model long-term individual preferences from past decisions,
(ii) adapt when preferences evolve with new context and constraints, and
(iii) resolve each conflict by explicitly grounding decisions in the inferred user priors. 

The explosive growth of LLMs has enabled the development of language agents. Their ability to perceive and reason over complex information shows promise as intelligent assistants that automate real-world tasks across different domains, such as software development, chart generation, film-making, and travel planning \citep{codact, li-etal-2025-metal, qian2025userrl, li2025echofoley}. Yet it remains unclear whether their performance is \emph{trustworthy} for \emph{calendar conflict resolution}, where small mistakes compound and mis-modeled preferences directly translate into costly time allocation errors. This motivates a central question: 
\begin{quote}
    \emph{\textbf{Can we trust LLMs to manage time?}} 
\end{quote}

To enable a systematic investigation of this problem, we introduce \BenchName{}, a benchmark for evaluating language agents on calendar event conflict resolution. \BenchName{} features synthetic users with diverse organizational roles and year-long calendars populated with carefully designed conflict scenarios. Conflict events are presented sequentially over time, and the agent receives feedback after each decision. This interactive setup closely mirrors real-world calendar management, where agents must infer and adapt to user preferences progressively through repeated interaction, rather than relying on fixed or one-shot instructions.
Our empirical results show that current LLMs struggle on this task with high error rates. These failures reveal a fundamental limitation: \textbf{LLM agents have a \emph{weak} ability to infer, retain, and refine preference-driven decision principles over long horizons.}


To address this gap, we propose \ModelName{} (\textbf{P}reference \textbf{E}volving \textbf{A}gent with \textbf{R}einforcement \textbf{L}earning), a reinforcement learning framework that trains language agents to \emph{infer} user preferences online and \emph{apply} them consistently over long-horizon calendar conflicts. \ModelName{} introduces a structured rollout with a persistent external memory, the \emph{Strategy Hub}, which stores a set of interpretable decision strategies (preference states) and is iteratively retrieved and updated at each round to capture newly revealed user priorities. To make preference learning explicit and stable, we optimize the agent with a curriculum-based reward, gradually shifting emphasis from preference inference in early rounds to preference-consistent decision making in later rounds. Experiment shows that \ModelName{} achieves an 0.76 error reduction rate on \BenchName{}, and 55\% improvement in average error rate compared to the strongest baseline.

In summary, our main contributions are:
\begin{itemize}[nosep, leftmargin=*]
    \item \textbf{Task.} We formulate \emph{calendar conflict resolution} as a new challenging task for LLMs agents, requiring preference-sensitive decision-making for conflict events over long horizons.
    \item \textbf{Benchmark.} We construct \BenchName{}, an evaluation suite with a synthetic data generation engine and standardized evaluation protocols to systematically evaluate LLM agents on calendar conflict resolution, and we provide an in-depth analysis of their failure modes.
    \item \textbf{Method.} We propose \ModelName{} (\S \ref{sec:method}), a reinforcement learning framework that enables agents to progressively infer and adapt to user preferences on-the-fly with an explicit memory module and carefully designed round-wise rewards, improving average error rate by 55\% over the strongest baseline  on \BenchName{}.
\end{itemize}

\begin{figure}[!t]
    \centering
    \includegraphics[width=1\linewidth]{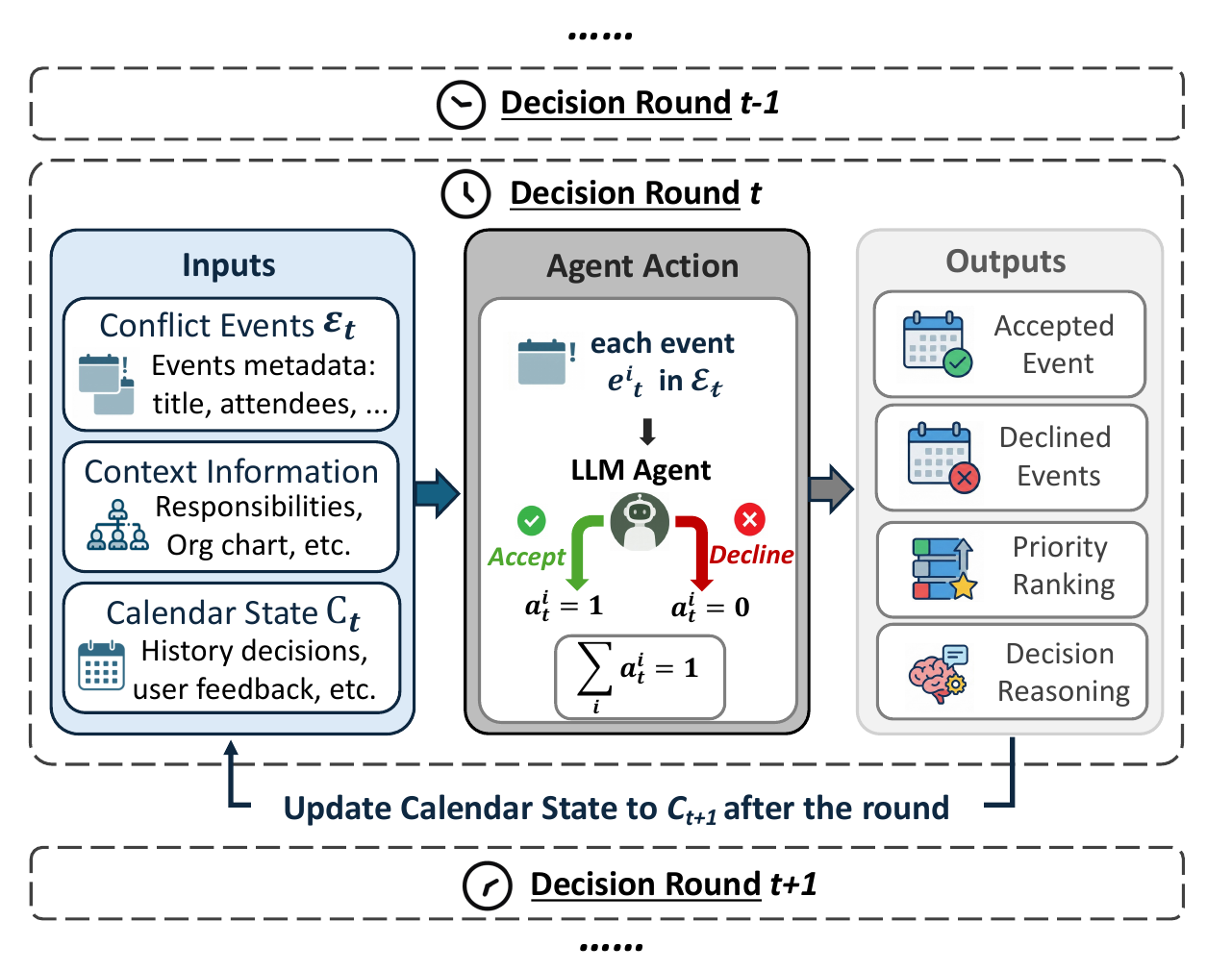}
    \vspace{-0.3in}
   \caption{\textbf{Illustration of the \textit{proposed} calendar conflict resolution task.} At decision round $t$, the agent observes (i) the conflicting events $\mathcal{E}_t$, (ii) contextual information, and (iii) the current calendar state $\mathcal{C}_t$. The agent selects exactly one event to accept ($a_t^i=1$) and declines the rest ($a_t^i=0$), producing the accepted event, declined events, a priority ranking, and rationale.}
\label{fig:task_illustration}
\end{figure}

\section{Task Formulation}
\label{sec:task_formulation}

In this section, we formally define the proposed \TaskName{} task. Appendix \ref{app:example_data} illustrates an example data point.

\noindent\textbf{Task Objective.}
The task is modeled as a sequential decision process with state transitions. As illustrated in Figure \ref{fig:task_illustration}, the goal of \TaskName{} is to construct a valid calendar for a single user by resolving a sequence of event conflicts over time. At each step $t$, the agent is presented with the current calendar state $\mathcal{C}_t$, and a set of temporally overlapping events $\mathcal{E}_t = \{e_t^1, \ldots, e_t^{N_t}\}$ and must accept exactly one event $e_t^i \in \mathcal{E}_t$, rejecting all others. The objective is to progressively model user preferences through interaction and contextual signals, producing a final calendar state $\mathcal{C}_T$ that aligns with the user’s preferences and decision context.

\noindent\textbf{Agent Action Space.}
At step $t$, the agent is tasked with assigning a binary decision $a_t^i \in \{0,1\}$ to each event $e_t^i \in \mathcal{E}_t$, where $a_t^i = 1$ denotes acceptance and $a_t^i = 0$ denotes rejection. The action must satisfy the constraint $\sum_i a_t^i = 1$.

\noindent\textbf{Environment Observation Space.}
The observation space is designed to reflect real-world calendar usage. At each step $t$, the agent observes contextual information (e.g. organization chart), the current calendar state $\mathcal{C}_t$ , and the set of conflicting events $\mathcal{E}_t$.
Each event $e_t^i \in \mathcal{E}_t$ is represented by structured metadata, including temporal attributes (e.g., start and end times), participant information, event descriptions (e.g. meeting topic or event summarization).
The calendar state $\mathcal{C}_t$ summarizes previous calendar events and user decisions. 


\section{CalConflictBench}
\label{sec:benchmark}

We introduce \BenchName{} to support the evaluation of the proposed task. In the benchmark, we present a synthetic data engine (Section~\ref{sec:data_engine}) for generating realistic, role-specific calendars and a comprehensive evaluation protocol (Section~\ref{sec:protocol}).

\subsection{Synthetic Data Engine}
\label{sec:data_engine}

We construct the synthetic data engine to generate data for training and evaluation. We report the details of data engine design in Appendix \ref{app:data_engine}, and we summarize key steps as follows.

\noindent\textbf{Organizational Schema Curation.}
We begin by crafting organizational schemas that capture real-world structures (e.g., research laboratories and technology companies). We conduct interviews with domain practitioners and analyze the collected real-world calendar data and organizational charts to extract role-specific information for each position (e.g. PI, postdoc, PhD student; CEO, SWE, HR).
For each role, we curate schemas based on the extracted information, including:
(1) regular meeting schemas, such as typical topics, frequencies, and attendees;
(2) priority principles $P$ that govern decision-making (e.g., leadership duties, deadline sensitivity, people management);
and (3) common conflict reasons $C$ (e.g., deadline clashes, hierarchical obligations, external commitments).
These priority principles are not directly observable by the agent. We further perform human verification on all schema to ensure reliability.

\noindent\textbf{Step 1: Synthetic Organization and User Profile Generation.}
Given an organizational schema, we instantiate user profiles for each role within the organization. Each user is associated with a fixed role, a regular meeting pattern, and a priority principle set. This step defines the ground-truth preference structure that governs all downstream calendar decisions.

\noindent\textbf{Step 2: Regular Event Generation.}
For each user, we generate a year-long calendar consisting of regular events using python scripts. Events are sampled according to role-specific meeting schemas, resulting in 52 weeks of weekly schedules. At this stage, calendars contain no conflicts and reflect the user’s normal workload and responsibilities.

\noindent\textbf{Step 3: Conflict Event Generation.}
We then carefully and systematically inject conflict events by overlapping regular events within the same time window. Given the user’s priority principles, conflict reasons, and predefined accept/decline ratios, we generate conflicting event sets together with a unique ground-truth resolution. These conflicts vary in difficulty, ranging from single-factor trade-offs to multi-factor conflicts that require balancing urgency, interpersonal relationships, and values.

\noindent\textbf{Step 4: Human Annotator Verification.} In the last step, we perform human verification to ensure the validity of the synthetic data and filter out implausible or inconsistent cases.

\subsection{Evaluation Protocol}
\label{sec:protocol}

Our evaluation is designed to assess the \emph{preference-evolving capability} of LLM agents, which is whether the agent can infer decision-making principles of users over time. Note that the evaluation is designed in a  \textbf{single-turn} format, and each instance contains history context (past-round information). 

\noindent\textbf{Parameters.}
We define three evaluation parameters:
(i) the total number of decision rounds $N$,
(ii) the context window size $W$, which specifies how many past rounds of information are provided to the agent, and
(iii) the total number of events are conflicting with each other per round $M$.

\noindent\textbf{Procedure.}
Each evaluation instance ( one trajectory ) simulates one year of calendar usage for a single synthetic user. Calendar conflicts are presented sequentially over time, mimicking realistic calendar dynamics. The agent does not have access to the ground-truth priority principles and must infer them solely from  history and contextual information. The agent may update its internal beliefs or strategies across rounds, and performance is evaluated over the full trajectory of $N$ rounds to capture long-horizon adaptation.

\noindent\textbf{Per-Round Metrics.}
We design the following metrics to evaluate decision quality at each round:

\begin{itemize}[nosep, leftmargin=*]
    \item \textbf{\textit{Decision Accuracy}}.
      A binary indicator of whether the agent’s accepted event matches the ground-truth accepted event. Note that invalid outputs are counted as incorrect.
    
    \item \textbf{\textit{Optimal Rank Distance (ORD)}}.
    For rounds with $M \geq 3$, we ask the agent to produce a ranking \(\rho_t\) over the \(M = |\mathcal{E}_t|\) candidate events. Let \(e_t^{*}\) be the ground-truth accepted event with 0-indexed position
    \(\mathrm{pos}_t(e_t^{*};\rho_t)\in\{0,\dots,M -1\}\). We define the Optimal Rank Distance ($ORD$) as
    \[
        ORD = 1 - \frac{ \mathrm{pos}_t(e_t^{*};\rho_t)}{M - 1}, \quad ORD \in [0,1].
    \]
    \vspace{-0.05in}
\end{itemize}

\noindent\textbf{Per-Instance Metrics.}
To measure preference learning and adaptation over time, we define three instance-level metrics:

\begin{itemize}[nosep, leftmargin=*]
    \item \textbf{\textit{Average Error Rate over $N$ rounds}}.
    The mean decision error across all $N$ rounds in a trajectory, capturing overall long-horizon performance.

    \item \textbf{\textit{Average ORD of $N$ rounds}}.
    The average ORD across all $N$ rounds in a trajectory, measuring how close the predicted event priority is to the optimal ranking.

    \item \textbf{\textit{Error Reduction Rate}}.
    The relative decrease in average error rate in the first quarter of the instance to average error rate in the last quarter of the same instance, measuring the agent’s ability to learn and improve its decisions over time.
\end{itemize}

\section{Evaluation}
\label{sec:experiments}

\begin{table*}[t]
\centering
\small
\setlength{\tabcolsep}{6.5pt}
\renewcommand{\arraystretch}{1.25}

\begin{tabular}{lccccccccccc}
\toprule
\multirow{2}{*}{}
& \multicolumn{5}{c}{\textbf{Average Error Rate of $N$ rounds}}
& \multicolumn{5}{c}{\textbf{Optimal Rank Distance of $N$ rounds }}
& \multirow{2}{*}{\makecell[c]{\textbf{Error}\\\textbf{Reduction}\\\textbf{Rate}}} \\
\cmidrule(lr){2-6}
\cmidrule(lr){7-11}
& $1$ & $25$ & $50$ & $75$ & $104$
& $1$ & $25$ & $50$ & $75$ & $104$
& \\
\midrule

\rowcolor{gray!8}
\multicolumn{12}{l}{\textit{\textbf{Base Models}}} \\

Qwen3-4B
&$0.44$ & $0.46$ & $0.44$ & $0.45$ & $0.45$
& $0.73$ & $0.73$ & $0.75$ & $0.75$ & $0.76$
& \ERR{-0.029}{} \\

Qwen3-8B
&$\mathbf{0.30}$ & $\underline{0.38}$ & $\underline{0.36}$ & $\underline{0.37}$ & $\underline{0.37}$
& $0.76$ & $0.78$ & $0.79$ & $0.79$ & $0.79$
& \ERR{0.026} \\

Qwen3-14B
&$0.38$ & $0.42$ & $0.41$ & $0.40$ & $0.41$
& $0.82$ & $0.75$ & $0.75$ & $0.74$ & $0.75$
& \ERR{-0.039} \\

Qwen3-30B
&$\underline{0.34}$ & $0.39$ & $0.39$ & $0.39$ & $0.38$
& $0.79$ & $\underline{0.79}$ & $0.79$ & $0.78$ & $0.78$
& \ERR{0.069} \\

Qwen3-30B-Think
&$0.36$ & $\underline{0.38}$ & $\mathbf{0.34}$ & $\mathbf{0.36}$ & $\mathbf{0.35}$
& $0.80$ & $\underline{0.79}$ & $\underline{0.81}$ & $\underline{0.81}$ & $\underline{0.82}$
& \textbf{\ERR{0.161}} \\

LLaMA-3.1-8B
&$0.66$ & $0.66$ & $0.67$ & $0.65$ & $0.65$
& $0.58$ & $0.58$ & $0.60$ & $0.61$ & $0.62$
& \ERR{-0.027} \\

OLMo3-7B-Instruct
&$0.98$ & $1.00$ & $1.00$ & $1.00$ & $1.00$
& $0.01$ & $0.00$ & $0.00$ & $0.00$ & $0.00$
& \ERR{-0.004} \\

OLMo3-32B-Think
&$0.40$ & $0.45$ & $0.46$ & $0.46$ & $0.45$
& $0.72$ & $0.72$ & $0.72$ & $0.72$ & $0.72$
& \ERR{0.050} \\

GPT-5-nano
&$\mathbf{0.30}$ & $0.42$ & $0.41$ & $0.43$ & $0.41$
& $\mathbf{0.85}$ & $0.77$ & $0.78$ & $0.77$ & $0.78$
& \underline{\ERR{0.122}} \\

GPT-5
&$0.42$ & $0.39$ & $\underline{0.36}$ & $\mathbf{0.36}$ & $\mathbf{0.35}$
& $0.83$ & $\mathbf{0.81}$ & $\mathbf{0.82}$ & $\mathbf{0.82}$ & $\mathbf{0.83}$
& \ERR{0.092} \\

Gemini-2.5-flash
&$\mathbf{0.30}$ & $0.40$ & $0.39$ & $0.40$ & $0.38$
& $\underline{0.84}$ & $\underline{0.79}$ & $0.79$ & $0.79$ & $0.81$
& \ERR{0.088} \\

\midrule

\rowcolor{gray!8}
\multicolumn{12}{l}{\textit{\textbf{Agentic Rollouts}}} \\

ReAct
&$\underline{0.34}$ & $0.40$ & $0.39$ & $0.39$ & $0.39$
& $0.78$ & $0.78$ & $0.79$ & $0.79$ & $0.80$
& \ERR{0.007} \\

Mem+ReAct
&$0.36$ & $\mathbf{0.37}$ & $0.39$ & $0.39$ & $0.40$
& $\underline{0.84}$ & $\mathbf{0.81}$ & $\underline{0.81}$ & $0.80$ & $0.79$
& \ERR{-0.162} \\

\bottomrule
\end{tabular}

\caption{\textbf{Performance across different numbers of rounds $N$.}
All results are evaluated with context window size $W=20$ and $M=5$ conflicting events per round. Results are averaged over ten independent instances. For each $N$, the best performance is shown in \textbf{bold}, and the second-best is \underline{underlined}.}
\label{tbl:main_results}
\end{table*}

\subsection{Setup.}
We follow the protocol described in Section~\ref{sec:protocol}. We vary $M \in \{2,3,4,5\}$ and $W \in \{1,5,10,20\}$ to control the combinatorial difficulty and historical context available at each decision round. More details are reported in Appendix \ref{app:eval}.

\noindent\textbf{Data.}
We evaluate agents on full-year calendars (52 weeks) constructed for ten synthetic users drawn from two synthetic organizations. To manage computational cost, we uniformly sample one decision round per week. Each evaluation trajectory therefore consists of 104 decisions (i.e. conflict events series), resulting in 1,040 total decisions.

\noindent\textbf{Models.}
We evaluate a diverse set of strong LLMs as agent base models, spanning open-source, reasoning-oriented, and proprietary families. Our open-source models include Qwen3-8B/14B/30B/30B-Think ~\cite{qwen3}, OLMo3-7B/OLMo3-32B-Think~\cite{olmo3}, and LLaMA-3.1-8B~\cite{llama3}. We also include GPT5-nano, GPT5~\cite{gpt5}, and Gemini-2.5-Flash~\cite{gemini} for proprietary model families. On top of these base models, we further evaluate representative agentic rollout style prompting, including ReAct~\cite{yao2023react} and Memory-Augmented ReAct~\cite{zhu2025llm}.

\subsection{Results and Analysis}

Table \ref{tbl:main_results} presents the evaluation results across different numbers of decision rounds $N$. We summarize key insights as follows.

\begin{figure}[h]
    \centering
    \includegraphics[width=1\linewidth]{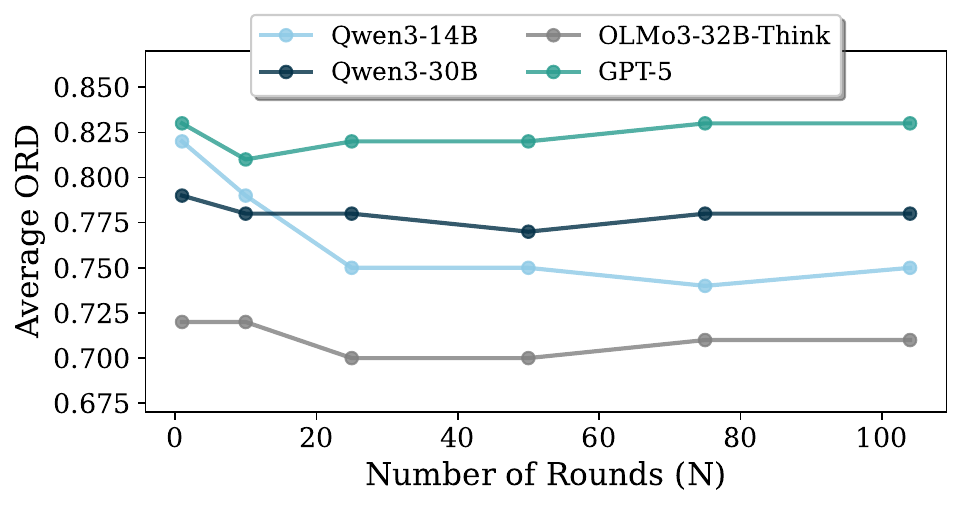}
    \caption{Average Optimal Rank Distance (ORD) over different numbers of decision rounds ($N$).}
    \vspace{-0.2in}
    \label{fig:ablation_n}
\end{figure}

\noindent\textbf{\textit{Insight 1}. Current LLMs do not exhibit Preference-Evolving capability.}
As indicated by the \emph{Error Reduction Rate} in Table \ref{tbl:main_results}, no evaluated LLM shows consistent performance improvement when transitioning from single-round ($N=1$) to multi-round settings. Error reduction rates are near zero or negative across models, including GPT-5 and Gemini-2.5-flash, suggesting that additional interaction rounds do not help refine decision principles. Figure \ref{fig:ablation_n} corroborates this finding, with error rates remaining flat or increasing as $N$ grows.

\begin{figure*}[t]
    \centering
    \vspace{0.1in}
    \begin{minipage}[t]{0.49\linewidth}
        \centering
        \includegraphics[width=\linewidth]{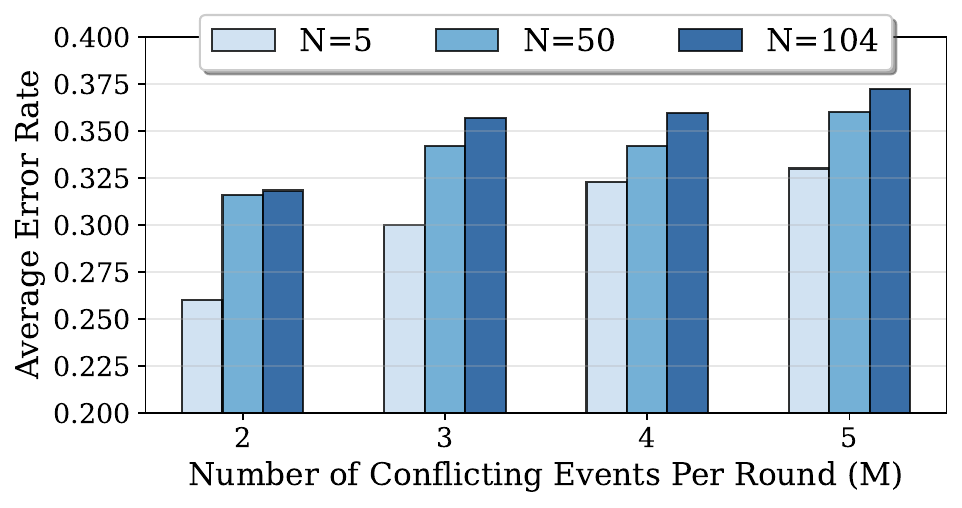}
    \end{minipage}\hfill
    \begin{minipage}[t]{0.49\linewidth}
        \centering
        \includegraphics[width=\linewidth]{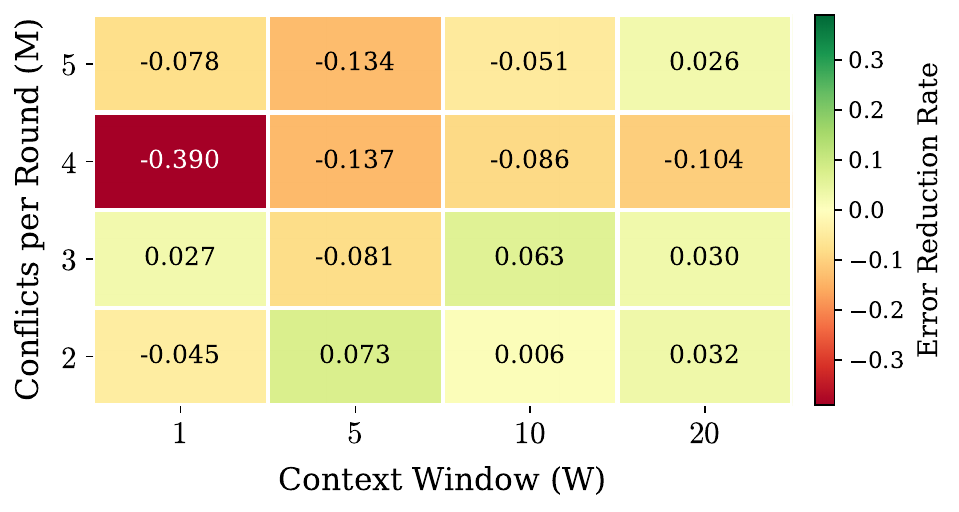}
    \end{minipage}
    \caption{Average Error Rate of Qwen3-8b under different numbers of conflicting events per round ($M$) (left), and Error Reduction Rate of Qwen3-8B under different evaluation parameters (right).}
    \vspace{-0.2in}
    \label{fig:ablation_n_w}
\end{figure*}

\noindent\textbf{\textit{Insight 2}. Increasing local decision complexity degrades performance.} As shown in Figure~\ref{fig:ablation_n_w} (left), the average error rate increases monotonically as the number of conflicting events per round $M$ grows. This trend reflects a rapid escalation in local decision complexity caused by higher event overlap, which expands the combinatorial decision space and increases ambiguity among candidate choices. Notably, this degradation is also observed in the single-round setting, indicating that errors arise primarily from local reasoning difficulty rather than long-horizon dependencies. As $M$ increases, these local errors accumulate across rounds, leading to compounded performance degradation in multi-round scenarios.

\noindent\textbf{\textit{Insight 3}. Larger context windows do not enable long-horizon reasoning.}
As shown in Figure \ref{fig:ablation_n_w} (right), increasing the context window size $W$ yields marginal and inconsistent changes in error reduction rate, with no clear monotonic improvement. In some cases, larger context windows even degrade performance, suggesting that additional context length  does not translate into better preference-aligned decisions, and it is insufficient for preference-evolving behavior.


\section{PEARL}
\label{sec:method}

\begin{figure*}
    \centering
    \includegraphics[width=1\linewidth]{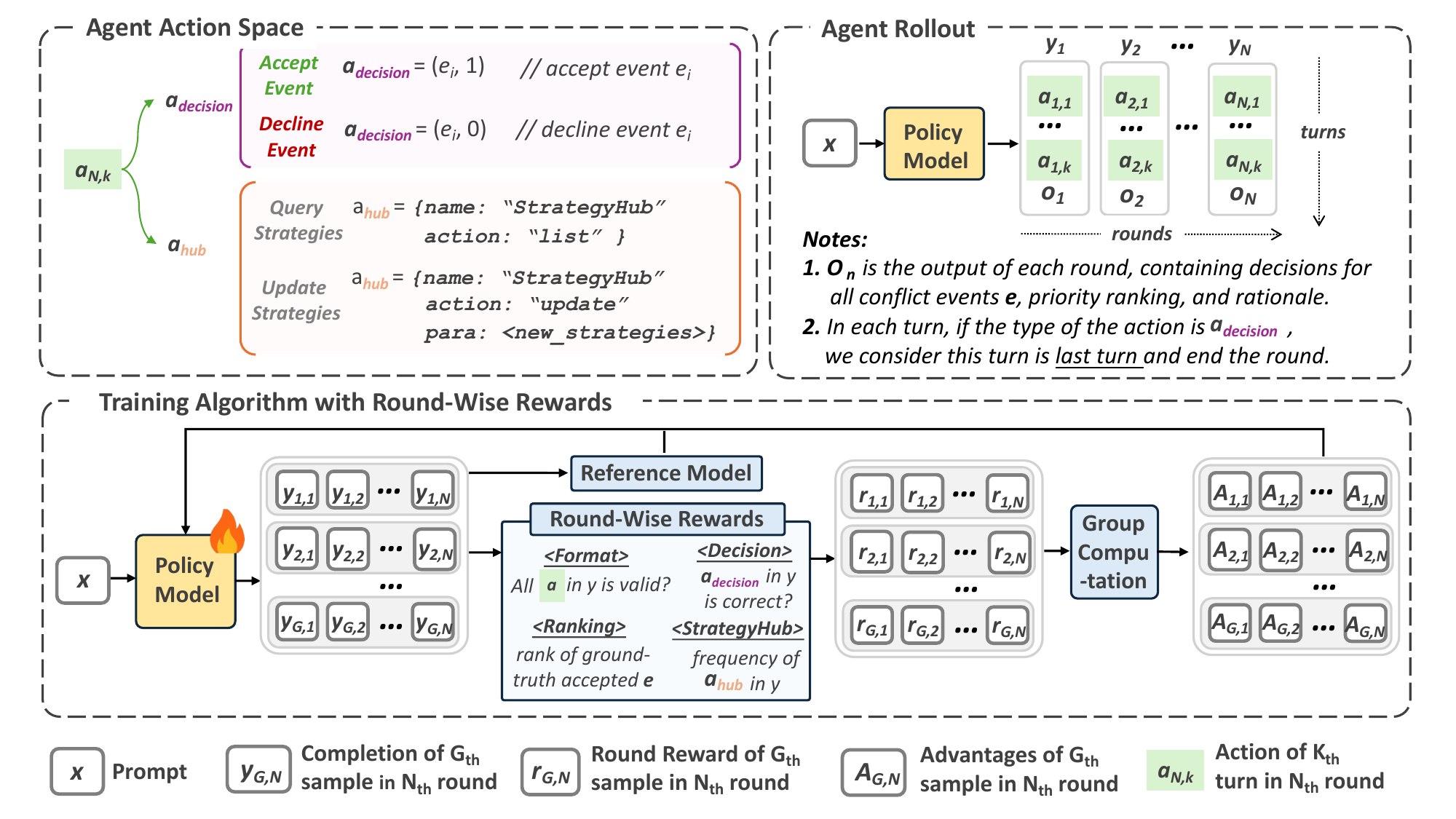}
    \caption{\textbf{Overview of \ModelName{}.} \textbf{Top-left: Agent action space.} At each turn, the agent can take a \emph{decision action} $a_{\text{decision}}$ (accept/decline an event $e_i$) or a \emph{hub action} $a_{\text{hub}}$ that queries (\texttt{list}) or updates (\texttt{update}) the external \emph{Strategy Hub}. \textbf{Top-right: Agent rollout.} The policy model generates a multi-turn trajectory; when a decision action is emitted, the round terminates and the next conflict is presented. \textbf{Bottom: Training with round-wise rewarding.} For each round, we sample multiple completions, score them with the curriculum-based reward model, and aggregate rewards into group-wise advantages by each round to update the policy.}
    \vspace{-0.12in}
    \label{fig:method}
\end{figure*}


We propose \ModelName{}, a reinforcement learning framework for long-horizon, preference-evolving language agents. In this section, we introduce our rollout design (Section~\ref{sec:strategy_hub}), reward modeling (Section~\ref{sec:reward}), and the experiment results for \ModelName{} evaluation (Section~\ref{sec:method_eval}).

\subsection{Rollout Design for Preference Inference}
\label{sec:strategy_hub}

We design a rollout mechanism that centers decision-making on a persistent, compact preference representation, enabling incremental inference and reuse across rounds.

\noindent\textbf{Strategy Hub.} Long-horizon preference learning via pure in-context history is challenging: As interactions grow, agents must repeatedly rediscover the same preference cues from a lengthy, noisy transcript, and the resulting preference state remains implicit and hard to reuse or update. To address this, we introduce the \emph{Strategy Hub} ($\mathcal{S}$) as an external memory module  that maintains a \emph{fixed-size} set of decision strategies. Each strategy encodes a user \emph{preference state} in natural language (See Appendix \ref{app:strategyhub} for details). The design of $\mathcal{S}$ explicitly separates \emph{preference inference}—identifying which strategy types matter and assigning their weights—from \emph{preference execution}—applying these learned priorities to new conflict contexts. This decomposition compresses preference learning into a compact and interpretable state that can be persistently updated across rounds, avoiding brittle reliance on implicit long-context representations.
 
At each decision round, the agent observes the current context (i.e. previous decisions and contextual information), and a set of conflicting events, and is granted access to $\mathcal{S}$, which is initialized as empty at the initial round. As shown in Algorithm~\ref{alg:strategy_hub_rollout}, the agent interacts with the $\mathcal{S}$ for a bounded number of turns (\(k\)), to retrieve and update strategies as needed.

\noindent\textbf{Agent Structured Rollout.} 
As illustrated in Figure \ref{fig:method}, the agent may take up to \(K\) turns within each round. At turn \(k\), it emits an action $a_{t,k} \in \mathcal{A} = \mathcal{A}_{\text{hub}} \cup \mathcal{A}_{\text{decision}}$, where \(\mathcal{A}_{\text{decision}}\) contains accept/decline decisions for events, and
\(\mathcal{A}_{\text{hub}}\) contains interactions with $\mathcal{S}$ (e.g., \texttt{list} current strategies or \texttt{update} current strategies).
We denote the round output as $O_t = (d_t,\rho_t,\xi_t)$, where
$d_t$ is the accept/decline decision set over events in $\mathcal{E}_t$ (typically accepting exactly one and declining the rest),
$\rho_t$ is the priority ranking over $\mathcal{E}_t$, and $\xi_t$ is the rationale.
The round terminates at the first turn $k_t$ such that $a_{t,k_t}\in\mathcal{A}_{\text{decision}}$.
The rollout can be written as a sequence of round outputs
\[
y = (O_1,\dots,O_N),
\qquad
\tau(x,y) = \{(o_t,O_t)\}_{t=1}^{N},
\]
Equivalently, the trajectory can also be represented by the per-turn action trace
$\{a_{t,k}\}_{t=1..N,\,k=1..k_t}$, where $k_t\le K$ is the stopping turn
when the decision action is emitted.

\begin{algorithm}[t]
\caption{\textbf{Agent Rollout Procedure}}
\label{alg:strategy_hub_rollout}
\small
\KwIn{StrategyHub $\mathcal{S}_0$; rounds $t=1..N$; context $\mathcal{C}_t$; conflicts $\mathcal{E}_t$; max turns $K$}
\KwOut{$y=(O_1,\dots,O_N)$, where $O_t=(d_t,\rho_t,\xi_t)$}

$\mathcal{S}\leftarrow \mathcal{S}_0$; $\mathcal{H}_{<1}\leftarrow \varnothing$\;
\For{$t\leftarrow 1$ \KwTo $N$}{
  $u_t\leftarrow 0$; $O_t\leftarrow \bot$\;
  $\mathcal{H}_{<t} \leftarrow \{\mathcal{C}_\tau^\star\}_{\tau < t}$
  \tcp*[r]{\textcolor{gray}{history}}
  \For{$k\leftarrow 1$ \KwTo $K$}{
    $a_{t,k}\sim \pi_\theta(\cdot\mid \mathcal{C}_t,\mathcal{H}_{<t},\mathcal{E}_t,\mathcal{S})$\;
    \uIf{$a_{t,k}\in\mathcal{A}_{\text{hub}}$}{
      \uIf{$a_{t,k}=\texttt{list}$}{\textsc{List}($\mathcal{S}$)\;}
      \ElseIf{$a_{t,k}=\texttt{update}(\Delta)$}{$\mathcal{S}\leftarrow \textsc{Update}(\mathcal{S},\Delta)$\;}
      $u_t\leftarrow 1$\;
    }
    \ElseIf{$a_{t,k}\in\mathcal{A}_{\text{decision}}$}{
      Parse $a_{t,k}$ into $(d_t,\rho_t,\xi_t)$; $O_t\leftarrow(d_t,\rho_t,\xi_t)$; \textbf{break}\;
    }
  }
}
\Return{$y$}
\vspace{-0.02in}
\end{algorithm}

\subsection{Reward Modeling for Preference-Evolving }
\label{sec:reward}

To train agents that both \emph{infer} user preferences and \emph{act} on them over long horizons, we design a curriculum-based reward model that encourages \emph{preference evolution} across rounds.

\noindent\textbf{Round-Level Rewards.} We assign rewards only at the round level. Each round \(t\) consists of up to \(K\) turns and terminates when the agent commits to a decision action or reaches the maximum number of turns \(K\). At each round \(t\), we design four reward signals that target complementary aspects at different granularities:
\begin{itemize}[itemsep=0.5pt, leftmargin=*]
    \item \textit{\textbf{Format Reward.}}
    To prevent catastrophic “invalid action” failures that break environment execution and learning, we reward outputs that are syntactically valid (i.e., parseable and in the allowed action space):
   $ r_t^{\text{f}}(x,y) \;=\; \mathbb{I}\!\left[a_t \in \mathcal{A}_{\text{valid}}\right]$.
   
    \item \textit{\textbf{Decision Reward.}} To directly optimize preference-aligned correctness, we reward the agent for making correct decision: $r_t^{\text{a}}(x,y) \;=\; \mathbb{I}\!\left[a_t = a_t^{*}\right]$, where \(a_t^{*}\) denotes the ground-truth round decision (accept / decline for events in \(\mathcal{E}_t\)).
    
    \item \textit{\textbf{Ranking Reward.}}
    To alleviate sparsity in \(r_t^{\text{a}}\), we add a denser signal based on the predicted priority ranking. We reward placing the ground-truth accepted event \(e_t^{*}\) closer to the top of the agent-produced ranking $\rho_t$ over the \(M = |\mathcal{E}_t|\) candidate events:     
    $r_t^{\text{r}}(x,y)
    \;=\; 1 - \frac{ \mathrm{pos}_t(e_t^{*};\rho_t)}{M - 1}.$
    
    \item \textit{\textbf{Strategy Hub Interaction Reward.}} To encourage deliberate preference retrieval/refinement rather than purely reactive decisions, we reward rounds where the agent performs a valid StrategyHub interaction (\(u_t\in\{0,1\}\)): $r_t^{\text{i}}(x,y) \;=\; u_t$.
\end{itemize}

\noindent\textbf{Trajectory-Level Curriculum.} In long-horizon calendar decisions, the agent faces a \emph{cold-start} problem: In early rounds, user preferences are poorly identified, so directly optimizing action correctness can be high-variance and brittle, while the most useful behavior is to \emph{extract and consolidate} preference evidence into persistent memory ($S$). As interaction progresses, the preference state becomes more stable; at that point, the learning signal should shift toward \emph{preference-consistent execution}, where fine-grained prioritization among many candidates matters.
To encourage this staged learning, we treat the format reward and decision reward weights,
$\lambda^{\text{f}}$ and $\lambda^{\text{a}}$, as fixed hyperparameters, and schedule the ranking reward and strategy hub interaction reward,
$\lambda^{\text{r}}$ and $\lambda^{\text{i}}$ weights, as a function of the round index. We define the normalized round index:
$
i_t \;=\; \frac{t}{N} \in [0,1].
$
Then, we set round-dependent weights by linear interpolation:
\[
\lambda_t^{\text{r}} = 0.5 * i_t,
\qquad
\lambda_t^{\text{i}} = 0.5 * (1 - i_t).
\]
The shaped per-round reward is
\[
\tilde r_t(x,y) =
\lambda^{\text{f}}\, r_t^{\text{f}}
+\lambda^{\text{a}}\, r_t^{\text{a}}
+\lambda_t^{\text{r}}\, r_t^{\text{r}}
+\lambda_t^{\text{i}}\, r_t^{\text{i}}
\]
\vspace{-0.1in}
and the trajectory return is computed as
\[
R(x,y) \;=\; \sum_{t=1}^{N}\gamma^{t-1}\,\tilde r_{t}(x,y).
\]



\noindent\textbf{Round-Wise Advantage Estimation.}The trajectory contains $N$ decision rounds, and the curriculum makes the reward distribution \emph{non-stationary across rounds}. If we normalize advantages using a single trajectory-level baseline, (i) later rounds can dominate the learning signal due to larger/more direct rewards, and (ii) early-round updates become noisy because their returns are intrinsically more uncertain (preferences are not yet identified).  To stabilize training and improve credit assignment, we further group the roll-outs based on the round position, and compute advantages \emph{separately for each round position}.
Let $\tilde r_{t,i}$ be the shaped reward of rollout $y_i$ at round $t$.
We compute a round-position return-to-go:
\[
\vspace{-0.1in}
G_{t,i}(x)\;=\;\sum_{\tau=t}^{N}\gamma^{\tau-t}\,\tilde r_{\tau,i}(x,y_i).
\]
For each round position $t$, we normalize these returns across the group:
\begin{align*}
\mu_t(x) &= \frac{1}{G}\sum_{i=1}^{G} G_{t,i}(x),\\
\sigma_t(x) &=
\sqrt{\frac{1}{G}\sum_{i=1}^{G}\big(G_{t,i}(x)-\mu_t(x)\big)^2+\varepsilon}.
\end{align*}
Then the round-wise advantages are
\[
\hat A_{t,i}(x,y_i)=\frac{G_{t,i}(x)-\mu_t(x)}{\sigma_t(x)}.
\]

\noindent\textbf{Objective.}
We train the policy with the standard clipped GRPO objective, adapted with our computed round-wise advantages $\hat A_{t,i}(x, y_i)$.

\subsection{Experiment}
\label{sec:method_eval}

\noindent\textbf{Setup.}
We adopt Qwen3-4B as the base language model. We compare \ModelName{} against three baselines under the same evaluation protocol as Section~\ref{sec:benchmark}: (i) \textbf{Zero-shot}, which directly prompts the base model to resolve conflicts; (ii) \textbf{Zero-shot + \textit{StrategyHub}}, which augments the prompt with access to the external Strategy Hub but without parameter updates; and (iii) \textbf{SFT}, which performs supervised fine-tuning on training data. Unless otherwise specified, all methods operate on the same observed context and interaction history at each round, and are evaluated over the same set of evaluation data as Section \ref{sec:experiments}. All training details are provided in Appendix \ref{app:training}.

\begin{figure}[bp]
    \centering
    \includegraphics[width=1\linewidth]{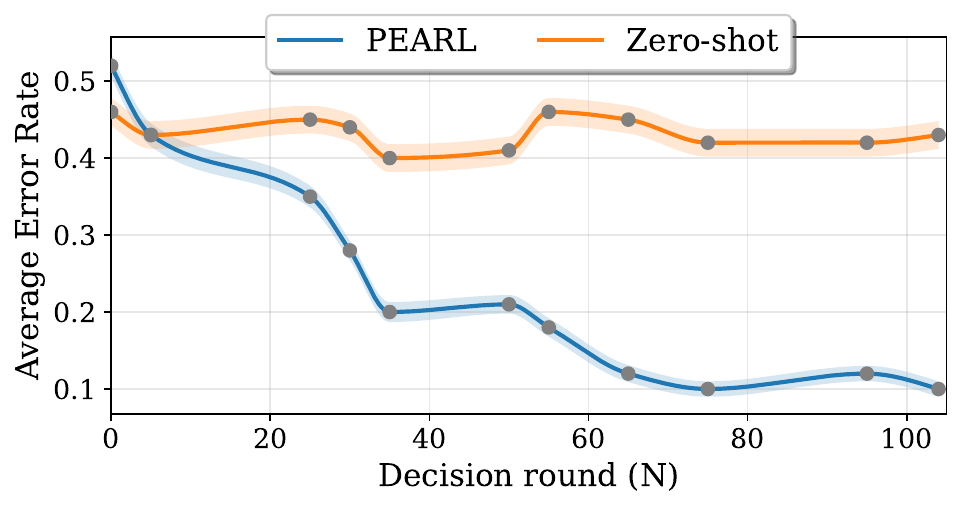}
    \caption{\textbf{Error vs. decision rounds} of \ModelName{} and zero-shot baseline}
    \label{fig:method_curve}
    \vspace{-0.2in}
\end{figure}


\noindent\textbf{Results and Analysis.}
Figure~\ref{fig:method_curve} reveals a clear separation in \emph{adaptation dynamics}. The zero-shot baseline stays nearly flat around a high error band across rounds, indicating that simply conditioning on growing history does not reliably improve preference alignment and can even slightly drift (negative ERR in Table~\ref{tbl:method}). In contrast, \ModelName{} exhibits a \emph{monotonic} reduction in error as the number of rounds $N$ increases, suggesting that it is not merely exploiting longer context, but is learning to \emph{update} its decision policy across decision rounds.

Table~\ref{tbl:method} further disentangles the sources of gains. Providing the memory module access alone (Zero-shot + StrategyHub) yields only modest improvement (\textit{\textbf{AER.}} decreases from $0.45$ to $0.41$; \textbf{\textit{ERR.}} increases from $-0.029$ to $0.048$), implying that \emph{having} an external memory without learning is insufficient for robust preference-evolving. Supervised training (SFT) improves final-round accuracy (with \textit{\textbf{AER.}} of $0.27$) but still lags behind \ModelName{} (with \textit{\textbf{AER.}} of $0.12$) and achieves substantially weaker adaptation (\textbf{\textit{ERR.}} $0.325$ vs.\ $0.761$). This gap suggests that imitation-style training learns better \emph{static} decision heuristics, yet struggles with long-horizon credit assignment and compounding preference-dependent errors across decision rounds over long horizon. Notably, \ModelName{} achieves a 55\% improvement in \textit{\textbf{AER.}} compared to the strongest baseline.

Overall, these results highlight that preference-evolving behavior requires \emph{long-horizon optimization} over multi-round trajectories: \ModelName{} can translate the history of previous rounds into measurable error reduction, validating the necessity of reinforcement learning for preference adaptation rather than one-shot prompting or purely SFT.

\begin{table}[tp]
\centering
\small
\setlength{\tabcolsep}{6pt}
\renewcommand{\arraystretch}{1.15}
\begin{tabular}{lcc}
\toprule
\textbf{Method} & \textbf{\textit{AER.} ($N$=104)}  & \textbf{\textit{ERR.}}  \\
\midrule
Zero-shot & $0.45$ & \ERR{-0.029} \\
SFT & $0.27$ & \ERR{0.325} \\
Zero-shot + \textbf{\textit{StrategyHub}} & $0.41$ & \ERR{0.048} \\
\midrule
\ModelName{} & $\textbf{0.12}$ ($\downarrow$) & \ERR{0.761} ($\uparrow$) \\
\bottomrule
\end{tabular}
\caption{\textbf{Final-round performance and adaptation.} Average Error Rate(\(\textbf{\textit{AER.}}\)) at the last decision round and Error Reduction Rate (\(\textbf{\textit{ERR.}}\)) across methods.}
\vspace{-0.2in}
\label{tbl:method}
\end{table}

\section{Related Work}
\label{sec:related_work}
LLM-based agents have been developed as intelligent assistants for tool-augmented question answering, web browsing, and real-world downstream tasks such as recipe generation and profile writing \citep{li-etal-2025-metal, qian2025userrl, lexical_planning}.Frameworks such as ReAct and AutoGPT enable autonomous behavior by interleaving reasoning and tool use \citep{yao2023react,yang2023autogpt}. Beyond tool-use, recent work casts LLM inference as an explicit planning/search problem, ranging from tree-based deliberation \citep{yao2023tree} and efficiency-oriented search \citep{katz2024thought} to interactive, code-augmented planners that execute and revise programs as plans \citep{liu2025interactive}. 
Complementary approaches learn planning-based reasoning by collecting trajectories and synthesizing process rewards for preference-based training \citep{jiao-etal-2024-learning}.
Yet personal time management remains less explored: earlier systems (e.g., Calendar.help) depended on predefined workflows with human-in-the-loop execution \citep{cranshaw2017calendar}; recent studies begin to investigate LLM-based scheduling agents \citep{shen2024smartcal,wijerathne2025scheduleme}. Our work extends this line to long-horizon calendar conflict resolution where agents must adapt to user-specific preferences over many decisions. Preference alignment is commonly achieved via RLHF, which fine-tunes models using human feedback \citep{ziegler2019finetuning,stiennon2020learning,ouyang2022training}; to reduce labeling cost, methods leverage AI-generated principles (e.g., Constitutional AI) \citep{bai2022constitutional}, and self-evaluation/self-correction \citep{wu2025selfplay}. Distinctly, we target preference alignment at test time under long horizons. Since long-horizon learning is hindered by limited context and state retention, prior work explores curriculum learning \citep{narvekar2020curriculum} and external memory/state tracking \citep{yan2025memoryr1}; we design external memory module to accumulate past decisions for preference inference and reuse across rounds.
\vspace{-0.1in}
\section{Conclusion}
\label{sec:conclusion}
In this work, we study calendar conflict resolution, a long-horizon, preference-driven decision-making task. We introduce \BenchName{} for systematic investigation, and evaluation results show that current LLM agents degrade as horizons grow and conflicts become denser. To address this, we propose \ModelName{}, a RL framework with an explicit memory module and round-wise rewards, achieving strong gains on \BenchName{}.
\section*{Limitations}
Our study is an initial step toward systematically evaluating and training preference-evolving agents for calendar conflict resolution, and it leaves several limitations for the future work. 
First, \BenchName{} represents user preferences via structured, role-conditioned rules over event attributes, which makes evaluation reproducible but inevitably incomplete. In real-world settings, decisions can be driven by transient and hard-to-observe factors that are not reflected in calendar metadata---e.g., ``I’m not in the mood for meetings today,'' fatigue, stress, interpersonal dynamics, or unexpected urgent tasks. Such affective and situational signals are difficult to simulate faithfully and may only be expressed through natural language messages or behavioral cues. Consequently, agents that perform well in our benchmark may still fail under implicit, rapidly shifting drivers of user choices.
Second, while we conduct all the necessary experiments to support our main claims, computational and time constraints prevent an exhaustive sweep over all possible combinations of evaluation parameters.
Third, because current LLMs have limited context windows, we only evaluate histories of up to 20 past events. We leave designing principled mechanisms for dynamically selecting and summarizing relevant context over long horizons as future work.

\section*{Acknowledgment}
This work is partially supported by supported by DARPA ITM Program No. FA8650-23-C-7316. The views and conclusions are those of the authors and do not necessarily reflect the official policy or position of the U.S. Government.

\bibliography{custom}

\appendix
\definecolor{codegreen}{rgb}{0,0.6,0}
\definecolor{codegray}{rgb}{0.5,0.5,0.5}
\definecolor{codepurple}{rgb}{0.58,0,0.82}
\definecolor{backcolour}{rgb}{0.95,0.95,0.92}
\lstdefinestyle{mystyle}{
    backgroundcolor=\color{backcolour},
    commentstyle=\color{codegreen},
    keywordstyle=\color{magenta},
    numberstyle=\tiny\color{codegray},
    stringstyle=\color{codepurple},
    basicstyle=\ttfamily\footnotesize,
    breakatwhitespace=false,
    breaklines=true,
    captionpos=b,
    keepspaces=true,
    numbersep=5pt,
    showspaces=false,
    showstringspaces=false,
    showtabs=false,
    tabsize=2
}
\lstset{style=mystyle}

\section*{Appendix}
\label{sec:appendix}

\section{Use of LLMs}
 In this work, LLMs are used strictly for research support rather than as sources of substantive content. Their use falls into: (i) serving as the tested and trained model, and (ii) assisting with language refinement during paper writing. For writing support, we used GPT-5 solely to polish text (improving coherence and grammar) while all ideas, logic, results, and technical contributions originate from the authors.

\section{Potential Risks}

Calendar conflict resolution is a high-stakes setting: incorrect accept/decline decisions can cause missed deadlines, lost opportunities, and interpersonal or organizational harm, especially over long horizons where errors compound. Calendar data and org context are also sensitive and can encode confidential relationships and priorities. Additionally, such agents could be misused for surveillance or coercive scheduling, and benchmark success may be over-interpreted because our setting models preferences as structured, role-conditioned rules that omit transient, hard-to-observe factors (fatigue, stress, interpersonal context). To mitigate these risks, we position our work as a controlled abstraction for reproducible evaluation rather than a deployment-ready system.

\section{Synthetic Data Engine Details}
\label{app:data_engine}

\subsection{Organizational Schema}
\label{app:schema}

Our synthetic data engine is grounded in \emph{role-conditioned organizational schemas} that capture how different positions operate and make trade-offs in calendar decisions. We first conduct semi-structured interviews with domain practitioners (e.g., PIs and PhD students in academia; executives and engineers roles in tech company) and analyze both (i) de-identified real-world calendar traces (event titles, recurrence patterns, attendee structures, meeting durations) and (ii) publicly available or provided organizational charts. From these sources, we extract role-specific attributes and encode them into a unified schema.

\paragraph{Schema fields.}
For each role $r$, we curate a schema $\mathcal{S}(r)$ consisting of three components:
\begin{enumerate}[leftmargin=14pt, topsep=2pt, itemsep=1pt]
    \item \textbf{Regular meeting schemas} $\mathcal{M}(r)$: templates for commonly recurring events, including (i) canonical
    topics (e.g., ``weekly group meeting'', ``1:1 mentoring'', ``sponsor sync''), (ii) typical cadence (weekly/biweekly/monthly),
    (iii) default duration distributions, (iv) attendee patterns (direct reports, cross-team stakeholders, external partners),
    and (v) common metadata realizations (location type, meeting modality, title variants).
    \item \textbf{Priority principles} $P(r)$: a small set of explicit, interpretable principles governing decisions under conflict,
    such as leadership/oversight obligations, deadline sensitivity, people management duties, and external relationship
    maintenance.
    \item \textbf{Conflict reasons} $C(r)$: common causes of decline/postpone for that role, such as deadline clashes, hierarchical
    obligations, travel constraints, task urgency spikes, teaching/committee constraints, or sponsor milestone collisions.
    Each conflict reason $c \in C(r)$ defines a transformation over event metadata (e.g., inserting a deadline marker, adding
    a senior attendee, changing modality to ``in-person required'').
\end{enumerate}

\paragraph{Unified representation.}
Concretely, a regular meeting template $m \in \mathcal{M}(r)$ is represented as
\[
m = \langle \texttt{topic},\; \texttt{freq},\; \texttt{dur},\; \texttt{attendees},\; \texttt{cts.}\rangle,
\]
where \texttt{constraints (cts.)} includes optional hard constraints (e.g., ``must be attended'', ``cannot be moved'') and soft constraints
(e.g., ``prefer mornings'', ``avoid back-to-back''). Priority principles are encoded as a weighted set
\[
P(r) = \{\langle p_k, w_k, g_k(\cdot)\rangle\}_{k=1}^{K_r},
\]
where $g_k(\cdot)$ is an attribute-based trigger function that maps an event (and local context) to $\{0,1\}$.
Conflict reasons are encoded as operators
\[
C(r) = \{\mathcal{T}_j\}_{j=1}^{J_r},
\]
where each $\mathcal{T}_j$ mutates an event into a plausible competing event (e.g., ``upgrade urgency'',
``attach deadline'').

\subsection{Conflict Event Generation}
\label{app:conflict_event_gen}

\begin{figure}[h]
    \centering
    \includegraphics[width=1\linewidth]{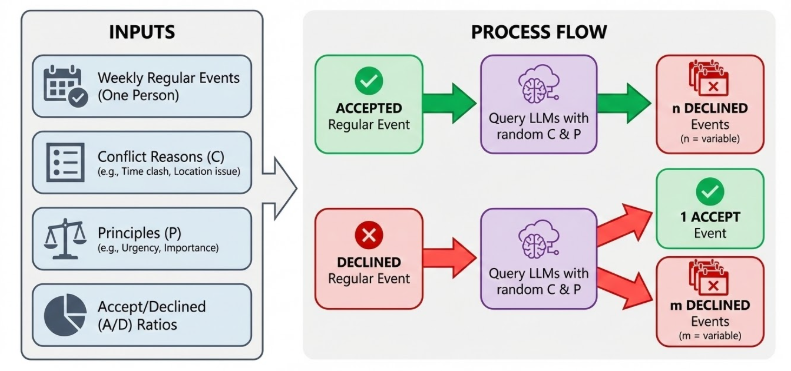}
    \caption{Conflict event generation process.}
    \label{fig:conflict:pipeline}
\end{figure}

Figure \ref{fig:conflict:pipeline} illustrated the conflict event generation process. Given a role-conditioned weekly calendar $\mathcal{C}$ sampled from $\mathcal{M}(r)$, we generate \emph{conflict rounds}
by constructing a candidate set of overlapping events $\mathcal{E}_t$ for each decision round $t$. Our generation procedure
explicitly couples each synthetic conflict with (i) a \emph{conflict reason} $c \in C(r)$ and (ii) a \emph{priority principle}
$p \in P(r)$ so that accepted/declined outcomes are explainable and consistent with role behavior.

\paragraph{Step 1: Sample anchor events.}
We first sample a set of \emph{anchor} regular events from the weekly calendar and assign each anchor a decision label
(accepted or declined) based on role-conditioned constraints and accept/reject ratio.
Intuitively, accepted anchors reflect high-priority routine obligations (e.g., weekly lab meeting for a PI), while declined anchors reflect lower-priority or optional events. The accept/decline ratio injects controlled randomness into the process.

\paragraph{Step 2: Generate competing events via principle--reason pairing.}
For each anchor event $e$ at round $t$, we sample a pairing $(p,c)$ where $p \sim P(r)$ (proportional to $w_p$ and triggers)
and $c \sim C(r)$, then apply the corresponding transformation to create competing events that overlap in time.
We denote the conflict generator as
\[
\mathcal{G}(e; p, c) \rightarrow \{e'_{1}, \dots, e'_{q}\},
\]
where each $e'_i$ inherits the timeslot of $e$ but differs in attributes (attendees, urgency, topic, location) induced by $(p,c)$.

\paragraph{Case A: accepted anchor $\rightarrow$ declined competitors.}
If the anchor $e$ is labeled \textbf{accepted}, we generate $n$ plausible \textbf{declined} competitors:
\[
\mathcal{E}_t \;=\; \{e\} \cup \{e'_1,\dots,e'_n\}.
\]
Competitors are created to be \emph{credible} yet dominated by $e$ under the role's principles, e.g.,
a PI's weekly group meeting competing with ad-hoc low-stakes chats.

\paragraph{Case B: declined anchor $\rightarrow$ one accepted competitor + extra declined.}
If the anchor $e$ is labeled \textbf{declined}, we generate (i) one \textbf{accepted} competitor $\hat e$ that is justified by a strong
principle trigger (e.g., deadline-driven sponsor call), plus (ii) $m$ additional \textbf{declined} competitors to increase local complexity:
\[
\mathcal{E}_t \;=\; \{\hat e\} \cup \{e\} \cup \{e'_1,\dots,e'_m\}.
\]
This construction ensures each round contains a non-trivial trade-off and supports ranking-based supervision: the accepted event
should be near the top even among multiple plausible alternatives.

\paragraph{Attribute realization and naturalization.}
To improve realism, we instantiate event surface forms using role-specific lexicons and title templates (e.g., ``1:1'', ``sync'',
``deep dive'', ``reading group'') and generate consistent metadata:
\begin{itemize}[leftmargin=14pt, topsep=2pt, itemsep=1pt]
    \item \textbf{Attendees:} sampled from the organizational chart with correct reporting lines (direct reports, peers, external partners).
    \item \textbf{Duration:} sampled from template distributions (e.g., 30min 1:1, 60min weekly meeting) with mild noise.
    \item \textbf{Urgency/deadlines:} inserted via $c$ (e.g., ``milestone due 5pm'', ``release cutoff today'').
    \item \textbf{Constraints:} hard constraints introduced for certain roles/events (e.g., committee meeting non-movable).
\end{itemize}

\subsection{Human Verification}
\label{app:human_verification}

We incorporate a human verification stage to ensure (i) \emph{plausibility} of event metadata, (ii) \emph{organizational consistency}
(attendee relations match the org chart), and (iii) \emph{decision validity} (accepted/declined labels align with the stated principles).
Annotators are provided with the role schema $\mathcal{S}(r)$, the organizational chart, and the conflict round $\mathcal{E}_t$,
and are asked to verify both the surface form and the underlying rationale.

\paragraph{Verification checklist.}
Each datapoint is reviewed with the following criteria:
\begin{enumerate}[leftmargin=14pt, topsep=2pt, itemsep=1pt]
    \item \textbf{Role realism:} Are the event topics and cadences plausible for this role?
    \item \textbf{Org-chart consistency:} Do attendees reflect correct reporting lines and stakeholder relationships?
    \item \textbf{Conflict coherence:} Do the competing events genuinely overlap and create a meaningful trade-off?
    \item \textbf{Principle alignment:} Is the accepted event justified by $P(r)$ under the provided context signals?
    \item \textbf{Metadata quality:} Are titles, locations, and constraints natural (no duplicates, no contradictions)?
\end{enumerate}

\paragraph{Edits and rejection.}
Annotators can (i) edit event titles/attributes, (ii) swap the accepted label if inconsistent with principles, (iii) rewrite the
conflict reason/context for coherence, or (iv) reject the datapoint if it cannot be repaired cheaply.

\paragraph{Annotation protocol.}
Each datapoint is reviewed by three annotators. The first two annotate independently, proposing edits and/or rejection decisions.
A third annotator then adjudicates disagreements and produces the final verified version by consolidating the two reviews. Data annotators are recruited from third party crowd-sourcing platform.

\subsection{Example Data}
\label{app:example_data}
Here is an example data point from generated synthetic organization.
\begin{tcolorbox}[
  enhanced,
  breakable,
  boxrule=0.8pt,
  arc=3pt,
  left=6pt,right=6pt,top=6pt,bottom=6pt,
  colback=white,colframe=black,
  title=\textbf{Example Datapoint: Input (Decision round $t$)},
  fonttitle=\small\bfseries
]
\small
\textbf{User.} PhD student (\textbf{James Carter}) at the \textbf{BioInnovate Research Lab}.\\[2pt]

\textbf{Conflict events $\mathcal{E}_t$ (overlapping timeslot on 2025-01-03).}
\begin{itemize}[leftmargin=12pt, topsep=2pt, itemsep=1pt]
  \item $e_1$: \textit{``Experiment planning and daily priorities sync''} \;\; 14:30--14:45 \;\; Attendees: J. Carter, E. White, M. Lee, S. Mitchell \;\; Type: internal coordination \;\; Location: BioInnovate Research Lab Conference Room
  \item $e_2$: \textit{``Equipment calibration check --- imaging suite''} \;\; 14:32--14:40 \;\; Attendees: J. Carter, E. White \;\; Type: operations/quality control \;\; Location: BioInnovate Research Lab Conference Room
  \item $e_3$: \textit{``Data preprocessing script optimization''} \;\; 14:35--14:42 \;\; Attendees: J. Carter, M. Lee, E. White \;\; Type: technical unblock \;\; Location: BioInnovate Research Lab Conference Room
  \item $e_4$: \textit{``Weekly lab reading group: methods paper''} \;\; 14:38--15:44 \;\; Attendees: J. Carter, S. Mitchell \;\; Type: reading/discussion \;\; Location: BioInnovate Research Lab Conference Room
  \item $e_5$: \textit{``Career development info session: resume workshop''} \;\; 14:30--15:33 \;\; Attendees: J. Carter, A. Patel \;\; Type: professional development \;\; Location: BioInnovate Research Lab Conference Room
\end{itemize}
\vspace{-2pt}

\textbf{Context information.}
\begin{itemize}[leftmargin=12pt, topsep=2pt, itemsep=1pt]
  \item Lab mission/direction: advancing biomedical research through innovative methodologies (\texttt{bioinnovate.org}).
  \item User responsibilities: develop thesis research, run experiments, analyze data, write papers/present, contribute to mentoring junior students, take courses.
  \item \textbf{Organization chart of BioInnovate Research Lab.}

  \begin{itemize}[leftmargin=12pt, topsep=1pt, itemsep=1pt]
    \item \textbf{Dr.\ Sarah Mitchell} --- \textit{Principal Investigator} (Management).\\
    Responsibilities: scientific vision/long-term strategy; secure funding; mentor/supervise all members; external representation.
    \item \textbf{Dr.\ Emily White} --- \textit{Postdoctoral Researcher} (Research). Supervisor: PI (Sarah Mitchell).\\
    Responsibilities: lead projects; mentor students; proposals/reports; write/present manuscripts and talks.
    \item \textbf{Dr.\ Michael Lee} --- \textit{Postdoctoral Researcher} (Research). Supervisor: PI (Sarah Mitchell).\\
    Responsibilities: lead projects; mentor students; proposals/reports; write/present manuscripts and talks.

    \item \textbf{Aisha Patel} --- \textit{PhD Student} (Research). Supervisor: PI (Sarah Mitchell).\\
    Responsibilities: thesis research; analysis/papers/presentations; mentor juniors; coursework.
    \item \textbf{James Carter} --- \textit{PhD Student} (Research). Supervisor: PI (Sarah Mitchell).\\
    Responsibilities: thesis research; analysis/papers/presentations; mentor juniors; coursework.
    \item \textbf{Lila Nguyen} --- \textit{PhD Student} (Research). Supervisor: PI (Sarah Mitchell).\\
    Responsibilities: thesis research; analysis/papers/presentations; mentor juniors; coursework.
    \item \textbf{Rajiv Sharma} --- \textit{PhD Student} (Research). Supervisor: PI (Sarah Mitchell).\\
    Responsibilities: thesis research; analysis/papers/presentations; mentor juniors; coursework.
    \item \textbf{Nina Garcia} --- \textit{PhD Student} (Research). Supervisor: PI (Sarah Mitchell).\\
    Responsibilities: thesis research; analysis/papers/presentations; mentor juniors; coursework.

    \item \textbf{Samuel Lee} --- \textit{Master's Student} (Research). Supervisor: PI (Sarah Mitchell).\\
    Responsibilities: focused research project; data collection/analysis/documentation; present results; coursework.
    \item \textbf{Mia Thompson} --- \textit{Master's Student} (Research). Supervisor: PI (Sarah Mitchell).\\
    Responsibilities: focused research project; data collection/analysis/documentation; present results; coursework.
    \item \textbf{Elena Martinez} --- \textit{Master's Student} (Research). Supervisor: PI (Sarah Mitchell).\\
    Responsibilities: focused research project; data collection/analysis/documentation; present results; coursework.
    \item \textbf{Noah Kim} --- \textit{Master's Student} (Research). Supervisor: PI (Sarah Mitchell).\\
    Responsibilities: focused research project; data collection/analysis/documentation; present results; coursework.
    \item \textbf{Olivia Rodriguez} --- \textit{Master's Student} (Research). Supervisor: PI (Sarah Mitchell).\\
    Responsibilities: focused research project; data collection/analysis/documentation; present results; coursework.

    \item \textbf{Jordan Rivera} --- \textit{Undergraduate Research Assistant} (Research). Supervisor: Postdoc (Emily White).\\
    Responsibilities: support experiments; data entry/basic analyses; attend lab meetings/reading groups; coursework.
    \item \textbf{Sophia Chen} --- \textit{Undergraduate Research Assistant} (Research). Supervisor: Postdoc (Emily White).\\
    Responsibilities: support experiments; data entry/basic analyses; attend lab meetings/reading groups; coursework.
    \item \textbf{Lucas White} --- \textit{Undergraduate Research Assistant} (Research). Supervisor: Postdoc (Emily White).\\
    Responsibilities: support experiments; data entry/basic analyses; attend lab meetings/reading groups; coursework.
    \item \textbf{Zoe Anderson} --- \textit{Undergraduate Research Assistant} (Research). Supervisor: PhD (Aisha Patel).\\
    Responsibilities: support experiments; data entry/basic analyses; attend lab meetings/reading groups; coursework.
    \item \textbf{Ethan Park} --- \textit{Undergraduate Research Assistant} (Research). Supervisor: PhD (Aisha Patel).\\
    Responsibilities: support experiments; data entry/basic analyses; attend lab meetings/reading groups; coursework.
  \end{itemize}
\end{itemize}
\vspace{-2pt}

\textbf{Calendar state $\mathcal{C}_t$.}
\begin{itemize}[leftmargin=12pt, topsep=2pt, itemsep=1pt]
  \item \textbf{Prior decisions.} (\textcolor{blue}{Note: The information is summarized here due to limited space, but we provided the full calendar information during evaluation and training}) On 2025-01-01 (14:15--15:45), user accepted \textit{``PhD qualifying exam planning session''} and declined: \textit{``Coursework and professional development check-in''}, \textit{``Weekly lab meeting -- planning and status update''}, \textit{``Internal project brainstorming session''}, and \textit{``Blue-sky reading discussion''}.
\end{itemize}
\end{tcolorbox}

\begin{tcolorbox}[
  enhanced,
  breakable,
  boxrule=0.8pt,
  arc=3pt,
  left=6pt,right=6pt,top=6pt,bottom=6pt,
  colback=white,colframe=black,
  title=\textbf{Example Datapoint: Output},
  fonttitle=\small\bfseries
]
\small
\textbf{Accepted event.} $e_2$ (Equipment calibration check --- imaging suite).\\[2pt]
\textbf{Declined events.} $e_1$ (daily priorities sync), $e_3$ (preprocessing script optimization), $e_4$ (reading group), $e_5$ (resume workshop).\\[4pt]

\textbf{Priority ranking $\pi_t$ (high $\rightarrow$ low).}
\[
\pi_t:\; e_2 \;>\; e_3 \;>\; e_1 \;>\; e_4 \;>\; e_5
\]
\vspace{-2pt}

\textbf{Decision reasoning.}
\begin{itemize}[leftmargin=12pt, topsep=2pt, itemsep=1pt]
  \item \textbf{Time-critical quality control:} calibration directly protects near-term experiment validity and data quality; missing it risks invalid data / wasted instrument time.
  \item \textbf{Execution before coordination:} once instruments are calibrated, follow-up coordination/unblocking can proceed with higher efficiency.
  \item \textbf{Unblocking is next:} preprocessing optimization resolves a pipeline blocker and likely accelerates the day’s progress, but is slightly more flexible than a calibration window.
  \item \textbf{Sync is helpful but deferrable:} the broader stand-up includes PI/postdocs, yet can be replaced by an async update if needed.
  \item \textbf{Deferable learning/career items:} reading group and resume workshop have lower immediate cost to reschedule and are not tied to a hard operational dependency.
  \item \textbf{History-consistent:} prior choices favored milestone-/execution-critical sessions over routine meetings and exploratory discussions.
  \item \textbf{Reschedule suggestion:} move $e_3$ to 14:45--15:00 for rapid follow-up; send a brief written status update to $e_1$ attendees.
\end{itemize}
\end{tcolorbox}

\section{Evaluation Details}
\label{app:eval}

We underscore again here that the evaluation in section \ref{sec:experiments} is conducted in single-turn manner.

\subsection{Prompt Template}
We attached the prompt template used for evaluation in section \ref{sec:experiments}.
\begin{lstlisting}[language=yaml]
prompt_template: |
    You are tasked with resolving a calendar conflict by analyzing the situation and making a decision based on organizational context and historical patterns.

    # Task: 
    1. Evaluate all conflict events considering:
       - The principles and reasoning provided for each event
       - The organizational hierarchy and relationships
       - The urgency and importance of each event
       - Historical patterns from similar past decisions
       - The impact on stakeholders and organizational goals
       - Time constraints and scheduling flexibility
    2. Rank all conflict events (including the regular event) in order of priority
    3. Select the single event that should be accepted
    4. Respone in the required format.

    # Inputs:

    ## History Conflict Calendar Events and User Decisions:
    {history_calendar_events}

    ## Organization Chart:
    {org_chart}

    ## Conflict Calendar Event to Solve:
    {conflict_calendar_event}

    #Output Format:
    Provide your response in the following structured format:
    ```json
    {{
        "priority_ranking (total {M} events)": ["ranked_event_id_1", ... , "ranked_event_id_{M}"],
        "reasoning": "Brief explanation of priority ranking and why the selected event was accepted",
        "selected_event_to_accept": "event_id"
    }}
    ```
\end{lstlisting}

\subsection{More Evaluation Settings}

Since our evaluation uses a single-turn interface, we implement agentic rollouts as a chain-of-thought-style output schema. For the ReAct baseline, we prepend a ReAct-style system prompt that instructs the model to produce an explicit \texttt{<reasoning>...</reasoning>} block followed by a \texttt{<response>...</response>} block. For ReAct + Memory, we additionally require a brief memory-aware analysis in an \texttt{<observation>} field: the model first emits \texttt{<observation>...</observation>} containing the provided past-round context, then generates \texttt{<reasoning>...</reasoning>}, and finally outputs \texttt{<response>...</response>}.

\subsection{Case Study}

\paragraph{Scenario.}
We analyze a representative conflict round where the agent must choose between a \textbf{doctor appointment} and an internal
\textbf{SEV2 incident meeting}. Both events overlap in time, and the user context indicates this is a \emph{personal healthcare}
commitment (non-delegable, often hard to reschedule) versus a mid-severity operational sync (important, but potentially
delegable and recoverable via async updates).

\begin{figure}[h]
    \centering
    \includegraphics[width=1\linewidth]{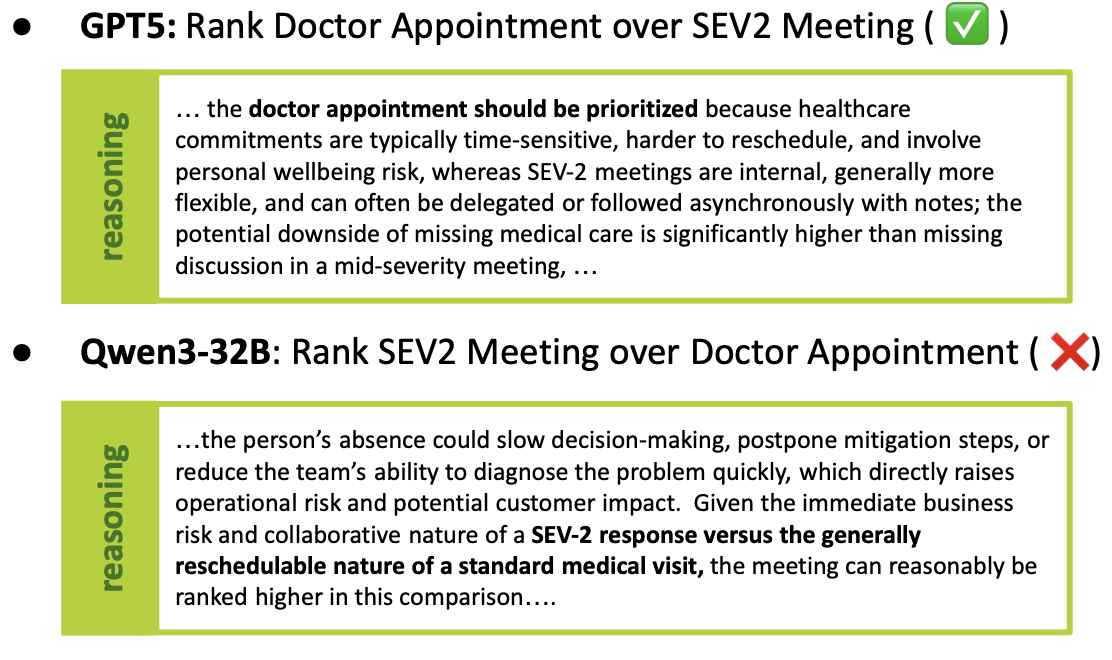}
    \caption{Case study: Responses from two models}
    \label{fig:case_study_medical_sev2}
\end{figure}

\paragraph{Model behaviors.}
Figure~\ref{fig:case_study_medical_sev2} contrasts two models.
\textbf{GPT-5} correctly ranks the \emph{doctor appointment} above the \emph{SEV2 meeting}, emphasizing that healthcare
appointments are typically time-sensitive, have higher personal risk, and are harder to reschedule than many internal meetings.
In contrast, \textbf{Qwen3-32B} incorrectly prioritizes the \emph{SEV2 meeting}, arguing that missing the meeting could slow
mitigation and increase business risk.

\paragraph{Why this matters.}
This failure mode is not merely a ``wrong preference''---it reflects a deeper modeling gap in \emph{role- and person-conditioned}
decision policies. In real workflows, users frequently treat certain personal commitments as \textbf{hard constraints}:
\emph{non-delegable}, \emph{high cost to cancel}, and \emph{limited reschedulability}. Meanwhile, even urgent workplace meetings
often admit mitigations: sending a delegate, joining partially, or catching up asynchronously via notes and incident logs.

\paragraph{Error analysis.}
The incorrect choice is driven by two systematic biases:
\vspace{-2pt}
\begin{itemize}[leftmargin=14pt, topsep=2pt, itemsep=1pt]
    \item \textbf{Overweighting organizational risk signals.} The model over-generalizes from ``incident response'' to a near-hard
    obligation, treating \texttt{SEV2} as always overriding other commitments, without calibrating severity or availability of substitutes.
    \item \textbf{Undermodeling non-delegability and rescheduling friction.} The model implicitly assumes a medical visit is easily movable
    (``generally reschedulable'') and ignores hidden costs: lead times, clinician schedules, cancellation fees, and health risks from delay.
\end{itemize}

\section{PEARL Details}

\subsection{StrategyHub Details}
\label{app:strategyhub}

\noindent\textbf{StrategyHub Tool.}
We implement \textsc{StrategyHub} as an external tool exposed to the agent via function calling.
At each round, the \textsc{StrategyHub} is reset to an empty list, and the agent may invoke the tool to \emph{read} or \emph{update} it, which is carried across decision rounds. Unless otherwise specified, the StrategyHub has a maximum capacity of 10 entries.

\noindent\textbf{Provided Tool Schema.}
To support consistent tool use, we provide the agent with a fixed metadata specification describing the StrategyHub schema, available fields, and constraints:
\begin{lstlisting}[language=python]
description = "Manage a short list of concise strategies. Actions: `list`, `update`"

# metadata
_json = {
            "type": "function",
            "function": {
                "name": self.name,
                "description": self.description,
                "parameters": {
                    "type": "object",
                    "properties": {
                        "action": {
                            "type": "string",
                            "enum": ["list", "update"],
                            "description": "Operation to run on the strategy list. If action is `list`, the response will be the current strategies. If action is `update`, the response will be the updated strategies.",
                        },
                        "strategies": {
                            "type": "array",
                            "items": {
                                "type": "string",
                                "description": "Strategy text to add or replace with (each strategy should be <=350 characters).",
                            },
                        },
                    },
                    "required": ["action"],
                },
            },
        }
\end{lstlisting}

\noindent\textbf{System Prompt.}
To ensure a fair comparison, we keep the task prompt unchanged during evaluation. To make the agent aware of the available tool, we prepend an additional system prompt, shown below:
\begin{lstlisting}[language=yaml]
system_prompt: |
    You are a calendar conflict resolution agent. 
    Think step-by-step, you can use the StrategyHub tool to help you and return the final answer strictly in the required JSON.

    StrategyHub tool:
      - You can list strategies with {"action": "list"}.
      - You can update strategies with {"action": "update", "strategies": ["strategy 1", "strategy 2", ...]}.
      - Keep strategies short (<=350 chars) and only a small set of the most useful ones.
      - Decide yourself whether an update is helpful (e.g., when no strategies exist or when a better summary is identified). If so, call the tool in tool_call fashion before producing the final answer.

    Before answering, you should first call the StrategyHub tool to get the latest strategies.
    Then you should analyze the history calandar events and see if the current strategies are helpful or need to be updated.
      - If the current strategies are not helpful or empty, you should update the strategy and update it to the StrategyHub with StrategyHub tool.
      - If the current strategies are helpful, you should use them to help you answer the question.
\end{lstlisting}

\subsection{Training and Validation Data Details}

We construct four synthetic organizations, each containing 10 users.
For every user, we synthesize a one-year calendar with realistic recurring meetings and injected conflict episodes, and then pool the calendars across all users and organizations to form the full dataset.
We randomly split the resulting dataset into training and validation sets using an 80/20 ratio. For training efficiency, we set the environment parameters \(W=M=5\).

\subsection{Baseline Details}

\noindent\textbf{Zero-shot.}
This is the first single-turn baseline. We use direct prompting under the same evaluation setting described in Section~\ref{app:eval}.

\noindent\textbf{SFT.}
This is the second single-turn baseline. We implement the SFT baseline using the LlamaFactory framework~\cite{zheng2024llamafactory}. 
Due to limited computational budget, we fine-tune the base model in a single-turn setting. The SFT baseline is trained on the same training subset as \ModelName{}. We format the training data as independent single-turn conversations, where each decision round is treated as a separate example. 
We keep the model’s thinking mode enabled throughout training.

\noindent\textbf{Zero-shot + StrategyHub.}
This is the multi-turn baseline. We add the same system prompt as \ModelName{} and grant the agent access to the \textsc{StrategyHub} tool, but do not apply any training.

\subsection{PEARL Training Details}
\label{app:training}
We implement the training recipe based on rLLMs framework \cite{rllm2025}. Note that we didn't perform any cold-start SFT. We directly train with original checkpoint. To stabilize preference learning and avoid cross-user leakage within an episode, we ensure that each episode contains events from exactly one user.
Since training on 104-step trajectories is both unstable and prohibitively long, we instead train the model on shorter-horizon instances by setting the number of decision rounds to \(N=20\) for the training subset, while keeping validation aligned with the full evaluation setting by using \(N=104\).





\noindent\textbf{Computation Resource.}
All training is conducted on 8$\times$ NVIDIA H100 GPUs (80GB memory per GPU). The training is consumed around 40 GPU hours.

\noindent\textbf{Training Hyperparameters}. Training hyperparameters and system configurations are summarized in Table~\ref{tab:training_params}.

\begin{table*}[htbp]
\centering
\small
\setlength{\tabcolsep}{7pt}
\renewcommand{\arraystretch}{1.15}
\begin{tabular}{lll}
\toprule
\textbf{Group} & \textbf{Parameter} & \textbf{Value} \\
\midrule
\multirow{2}{*}{Algorithm}
& Advantage estimator & \texttt{algorithm.adv\_estimator=grpo} \\
& KL coefficient & \texttt{algorithm.kl\_ctrl.kl\_coef=0.001} \\
\midrule
\multirow{6}{*}{Model / PPO}
& Base model & \texttt{Qwen/Qwen3-4B} \\
& Learning rate & \texttt{actor\_rollout\_ref.actor.optim.lr=1e-6} \\
& PPO clip (high) & \texttt{actor\_rollout\_ref.actor.clip\_ratio\_high=0.28} \\
& Loss aggregation & \texttt{seq-mean-token-mean} \\
& Use KL loss term & \texttt{actor\_rollout\_ref.actor.use\_kl\_loss=False} \\
\midrule
\multirow{3}{*}{Batch / Length}
& Train batch size & \texttt{data.train\_batch\_size=16} \\
& Val batch size & \texttt{data.val\_batch\_size=10} \\
& Max prompt/response length & \texttt{16384 / 16384} \\
\midrule
\multirow{6}{*}{Rollout (train / val)}
& Rollout engine & \texttt{vllm} (\texttt{mode=async}) \\
& Samples per prompt (train) & \texttt{actor\_rollout\_ref.rollout.n=8} \\
& Temperature (train) & \texttt{0.7} \\
& Samples per prompt (val) & \texttt{actor\_rollout\_ref.rollout.val\_kwargs.n=1} \\
& Temperature (val) & \texttt{0.6} \\
& Top-p (val) & \texttt{0.95} \\
\midrule
Efficiency / Systems
& GPUs $\times$ nodes & \texttt{trainer.n\_gpus\_per\_node=8}, \texttt{trainer.nnodes=1} \\
& Max tokens per GPU (PPO) & \texttt{actor\_rollout\_ref.actor.ppo\_max\_token\_len\_per\_gpu=32768} \\
& vLLM GPU mem util. & \texttt{actor\_rollout\_ref.rollout.gpu\_memory\_utilization=0.85} \\
& Grad checkpointing & \texttt{actor\_rollout\_ref.model.enable\_gradient\_checkpointing=True} \\
\midrule
\multirow{2}{*}{Stepwise advantage}
& Enable & \texttt{rllm.stepwise\_advantage.enable=True} \\
& Mode & \texttt{rllm.stepwise\_advantage.mode=per\_step} \\
\bottomrule
\end{tabular}
\vspace{-6pt}
\caption{\textbf{Key training and rollout hyperparameters for \ModelName{} (Qwen3-4B).}}
\label{tab:training_params}
\end{table*}

\end{document}